\documentclass[conference]{IEEEtran}
\usepackage[utf8]{inputenc}
\usepackage[a4paper, right=0.6in, top=1in, bottom=1.21in, left=0.5in]{geometry}
\IEEEoverridecommandlockouts
\usepackage{cite}
\usepackage{amsmath,amssymb,amsfonts}
\usepackage{algorithmic}
\usepackage{graphicx}
\usepackage{textcomp}
\usepackage{xcolor}
\usepackage{booktabs}

\usepackage{subcaption}
\usepackage{url}

\usepackage{hyperref}
\usepackage{cite}
\usepackage{fancyvrb}
\usepackage{amssymb}
\usepackage{graphicx}
\usepackage{array}
\usepackage{subcaption}
\usepackage{float}
\usepackage{multirow}
\usepackage{tabularx}
\usepackage{longtable}
\usepackage{multicol}
\usepackage{listings}
\usepackage{verbatim}

\usepackage{tikz}
\usepackage{pgfplots}
\usepackage{pgfplotstable}
\pgfplotsset{compat=1.7}
\usepackage{subcaption}

\def\BibTeX{{\rm B\kern-.05em{\sc i\kern-.025em b}\kern-.08em
    T\kern-.1667em\lower.7ex\hbox{E}\kern-.125emX}}

\begin{document}
%

\title{Development of a Humanoid Robot Prototype for Multimodal Human-Robot Interaction}

\author{\IEEEauthorblockN{\textsuperscript{1}Thang Tran Viet, \textsuperscript{2}Thanh Nguyen Canh, \textsuperscript{1}Huy Uong Gia,  \textsuperscript{1}Phuc Dinh Van,\textsuperscript{1}Son Tran Duc, \\ \textsuperscript{1}Ngoc Minh Do and \textsuperscript{1}Xiem HoangVan$^*$}
\IEEEauthorblockA{
\textit{\textsuperscript{1} University of Engineering and Technology, Vietnam National University, Hanoi, Vietnam}\\
\textit{\textsuperscript{2} School of Information Science, Japan Advanced Institute of Science and Technology} \\
$^*$\textit{Corresponding authors  (xiemhoang@vnu.edu.vn)}
}}

\maketitle              

\begin{abstract}
Human-robot interaction (HRI) enables intuitive and intelligent collaboration between humans and robots in real-world environments. This paper introduces a humanoid robot prototype designed as a flexible testbed for developing and integrating artificial intelligence (AI) modules in HRI tasks. The system features a $12$ degree-of-freedom (DOFs) dual-arm mechanism and a $2$ DOFs head with an expressive LCD screen to express facial emotions. All hardware components are controlled by a custom-designed controller board with real-time AI processing supported by an onboard Jetson module. The system incorporates three AI modules: (1) gesture recognition using MediaPipe Pose and an LSTM classifier, (2) object detection with YOLO and 3D localization, and (3) voice-command processing through speech recognition and large language model(LLM)-based semantic parsing. The platform is validated through experiments on positioning accuracy, with results showing average manipulation errors of approximately $1.83$ cm. To demonstrate its versatility, experimental results show over $90\%$ task accuracy, with gesture recognition reaching $96\%$, speech recognition reaching $92\%$. The results confirm the effectiveness of the proposed system as a reproducible and accessible humanoid platform for research and prototyping in HRI.
\end{abstract}


\section{Introduction}

Humanoid robots are a subject of growing research interest due to their ability to interact naturally with people in human-centric environments, unlike traditional industrial robots. As invaluable platforms for education and research, they are designed to operate in dynamic, unstructured human-centric environments, offering natural interfaces for communication through speech, gestures, and facial expressions~\cite{zhang2023large, jahanmahin2022human}. Building on this foundation, the rapid advancement of artificial intelligence (AI), particularly in computer vision and natural language processing, has further reinforced the role of human–robot interaction (HRI) as a cornerstone of modern robotics, enabling machines to perceive, interpret, and respond to human behavior in increasingly collaborative scenarios~\cite{sharkawy2021survey, selvaggio2021autonomy}.

In recent years, several high-profile humanoid platforms have advanced the state-of-the-art in locomotion, manipulation, and social expression. Unitree Robotics’ humanoid series, including G1~\cite{unitreeG1} and newer R1~\cite{unitreeH1}, represents significant improvements in dynamic stability, energy-efficient locomotion, and dual-arm manipulation within compact form factors. These robots integrated advanced control systems, highly optimized mechanical architectures, and cutting-edge sensory technologies to attain superior performance. G1 focuses on enhanced sensing and superior balance for real-world environments, whereas R1 emphasizes affordability, customizability, and accessibility for research and education. Ameca, developed by Engineered Arts~\cite{arts2022ameca}, possesses the ability to express facial expressions and extremely expressive conversational abilities. This makes it one of the most human-like communication platforms to date~\cite{berns2024you}. In contrast, Tesla Optimus Gen2~\cite{Optimus} focuses on lightweight design, full-body mobility, and dexterous hands for delicate operations. Boston Dynamics’ Atlas~\cite{nelson2017petman} has demonstrated unprecedented agility with parkour-like performance. These systems illustrate a trend toward high-performance humanoid robots that integrate mobility, perception, and interaction for human-centered applications. 
Despite their technical sophistication, these platforms remain largely closed-source and prohibitively expensive, limiting their accessibility for education and academic research. Extended repair cycles, high manufacturing costs, and limited modularity make them less suitable for iterative prototyping or AI experimentation. This has motivated the development of affordable, open, and customizable humanoid robot platforms to democratize HRI research.

To address this challenge, several open-source educational platforms have emerged. NimbRo-OP2X~\cite{ficht2020nimbro} is an open-source humanoid robot designed by the University of Bonn. With a height of $135$cm and $19$kg, it features torque-controlled actuators, a GPU-enabled onboard computer, and modular 3D-printed components. Its affordability and advanced locomotion, perception, and manipulation capabilities make it attractive for both research and competition, though its relatively limited degrees of freedom restrict highly flexible tasks. The ROBOTIS OP2/OP3 series~\cite{roboticsop3}, by contrast, are compact, lightweight, and ROS-compatible systems that provide an accessible entry point to vision and control experiments. However, their low payload, limited actuation strength, and small form factor hinder applications in complex manipulation or naturalistic HRI. The PoppyHumanoid\cite{lapeyre2014poppy} emphasizes modularity and reproducibility using 3D-printed parts and Dynamixel motors, making it ideal for research and classroom settings. Overall, while these systems lower barriers to entry, they remain underpowered in terms of strength, embodiment, and versatility, leaving a gap for platforms that balance affordability with the capability to support sophisticated HRI research.

In parallel, recent works have explored the integration of modern AI into humanoid robotics. Several studies~\cite{liu2023llm, canhcontext2025} proposed a framework for integrating large language models (LLMs) to enable robots to execute manipulation tasks based on natural language commands. Zhang \textit{et al.}\cite{zhang2018voice} developed a dual-arm ROS-based robot that supports voice-controlled manipulation, focusing on real-time coordination. In gesture-based HRI, some methods~\cite{shen2020understanding, peral2022efficient, canhefficient2025} combined pose estimation and deep learning for real-time non-verbal action recognition. However, many of these approaches lack full-body physical embodiment or are limited to simulation environments, reducing their relevance for real-world deployment and evaluation.

To bridge the gap between high-performance yet inaccessible systems and low-cost but underpowered alternatives, we present a fully operational humanoid robot prototype tailored for academic research and HRI experimentation. Our platform combines affordability, modularity, and real-time AI integration across speech, vision, and gesture modalities, and is designed for deployment in real-world, human-facing scenarios. The main contributions of this work are:
\begin{enumerate}
\item Development of a fully 3D-printed, low-cost humanoid robot featuring a 12 DOFs dual-arm mechanism, a 2 DOFs head, and a 7 in LCD screen for expressive facial interaction. The system is built on ROS and uses an onboard GPU-enabled Jetson for AI inference.
\item Incorporation of AI modules for gesture recognition, object detection with stereo-based 3D localization, and natural language understanding through speech recognition and LLM-based semantic parsing.
\item Implementation and evaluation of two real-world HRI scenarios—gesture-based interaction and voice-command-driven object manipulation—with quantitative analysis showing high accuracy, low latency, and robust task execution.
\end{enumerate}

The remainder of this paper is structured as follows: Section~\ref{sec:method} details the robot’s hardware and software architecture; Section~\ref{sec:results} presents the experimental setup and performance evaluation; and Section~\ref{sec:conclusion} concludes with a discussion on limitations and future directions.

\begin{figure}[!ht]
\centering
\includegraphics[width=0.48\textwidth]{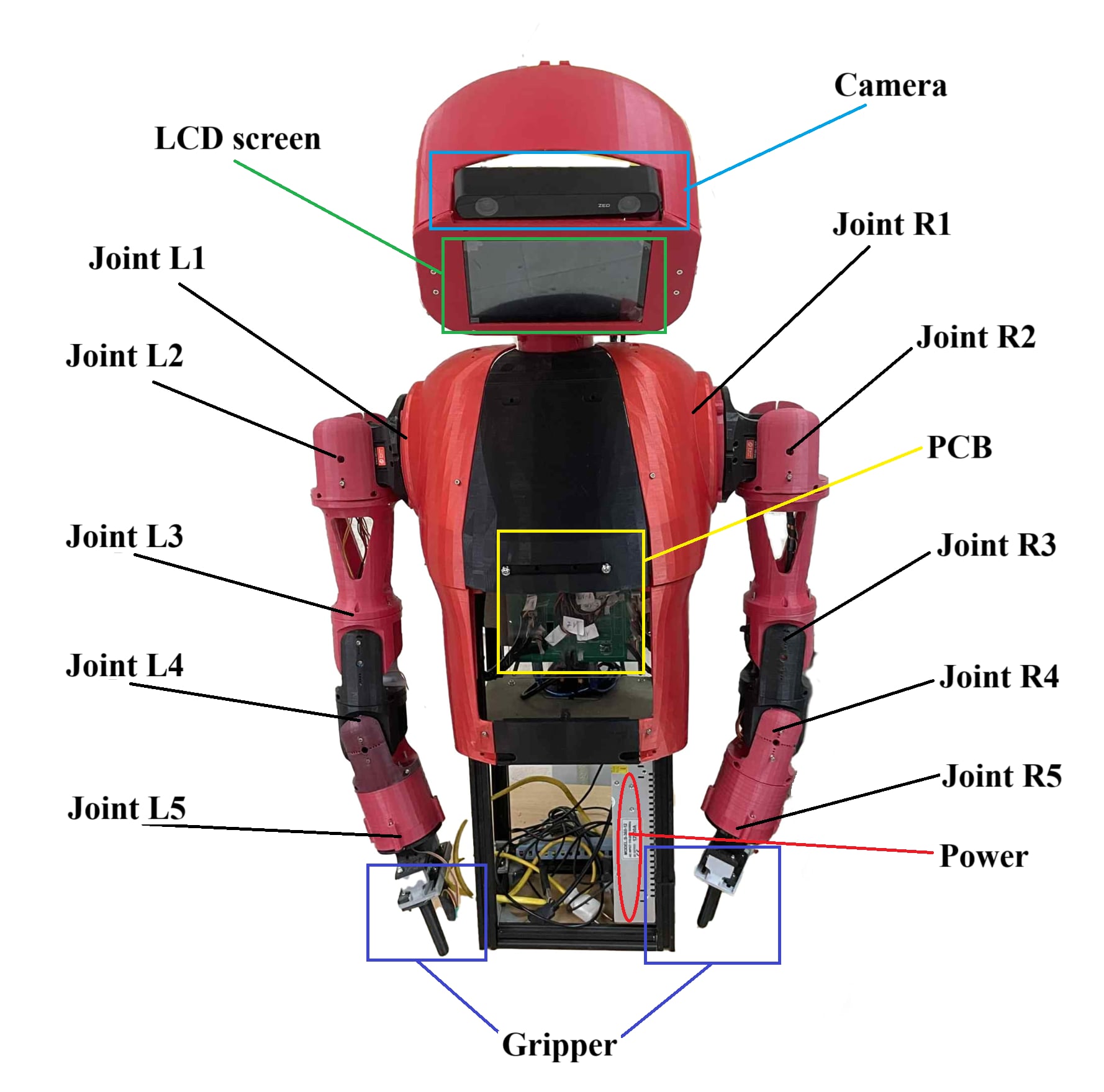}
\caption{Illustration of our hardware design.}
\label{fig:hardware}
\end{figure}

\section{Methodology} \label{sec:method}
This section details the design and development methodology of the proposed humanoid robot prototype, serving as a testbed for evaluating AI-based HRI algorithms. The system architecture is organized into two interconnected layers: the mechanical–electronic infrastructure, responsible for actuation and sensing, and the AI-powered perception and control layer, which enables responsive and adaptive behaviors.

\subsection{Mechanical and Electronic Structure}

The physical design of the humanoid robot is shown in Fig.~\ref{fig:hardware}. The robot has a modular architecture, integrating lightweight aluminum bars to build a frame and custom 3D-printed joints and brackets.  Aluminum provides durability and load-bearing capacity for the arms and body, while 3D-printed parts enable rapid fabrication, low-cost replacement, and customization of mechanical modules. This hybrid approach achieves both structural strength and ease of prototyping.  As a result, the prototype weighs only $15$ kg while approximately $150$ cm tall, which is close to the scale of an average human torso. This human-scale design enhances realism in HRI and enables gestures such as handshakes, waves, and object handovers to occur at natural heights, promoting intuitive experiments and their applicability in real-world contexts.

The robot features a 12-degree-of-freedom (DOF) dual-arm system, each arm composed of 6 revolute joints actuated by serial-link servo motors. The arm kinematics follow the standard Denavit–Hartenberg (D–H) convention for coordinate frame assignment and transformation, allowing both forward and inverse kinematics computation for manipulation tasks, which is detailed in our previous study~\cite{canh2024optimal}. The arms are designed to support pick-and-place operations, expressive gestures, and basic physical interactions. Table~\ref{tab:servo_specs} summarizes the key specifications of the servomotors used. The servomotors are allocated according to joint load requirements: the high-torque DS5160 units actuate the shoulders, FT5330M drives the elbows, and lighter TD-8120MG and MG996R units control the wrist and end-effector. This allocation balances strength and cost, achieving a maximum payload of approximately $600$g.

\begin{table}[!ht]
\centering
\caption{Main specifications of servomotors used in the humanoid robot}
\label{tab:servo_specs}
\begin{tabularx}{\linewidth}{  | >{\centering\arraybackslash}m{0.1\textwidth} 
  | >{\centering\arraybackslash}m{0.07\textwidth} 
  | >{\centering\arraybackslash}m{0.1\textwidth}  
  | >{\centering\arraybackslash}X|  }
\toprule
\textbf{Servo} & \textbf{Weight} & \textbf{Voltage Range} & \textbf{Max. Torque} \\
\midrule
DS5160 SSG   & 158\,g & 6.0–8.4\,V  & 65.0\,kg·cm \\
FT5330M      & 67\,g  & 6.0–7.4\,V  & 35.5\,kg·cm \\
TD-8120MG    & 56–68\,g & 4.8–8.4\,V & 22.8\,kg·cm \\
MG996R       & 55\,g  & 4.8–7.2\,V & 11.0\,kg·cm \\
\bottomrule
\end{tabularx}
\end{table}

\begin{figure}[!ht]
\centering
\begin{subfigure}[b]{0.32\textwidth}
\includegraphics[width=\textwidth]{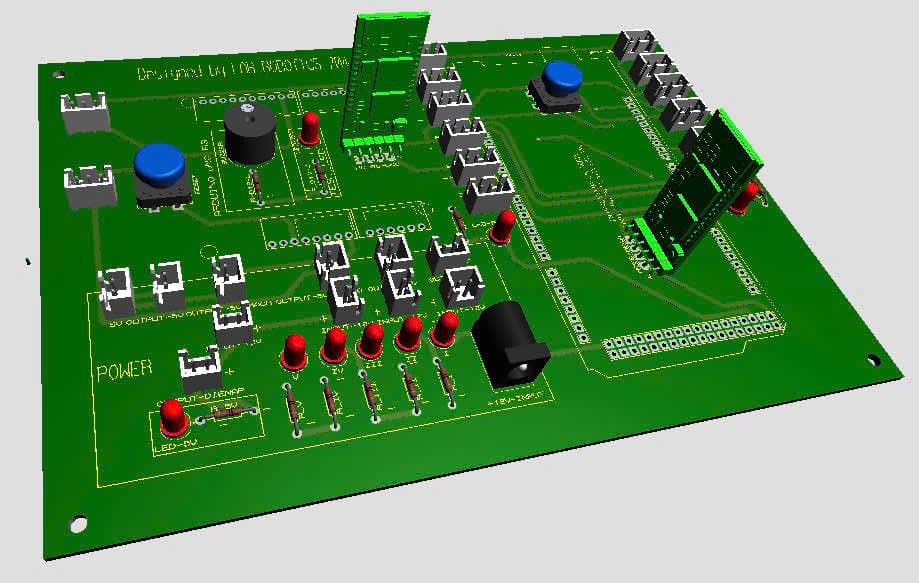}
\end{subfigure}
\hfill
\begin{subfigure}[b]{0.35\textwidth}
\includegraphics[width=\textwidth]{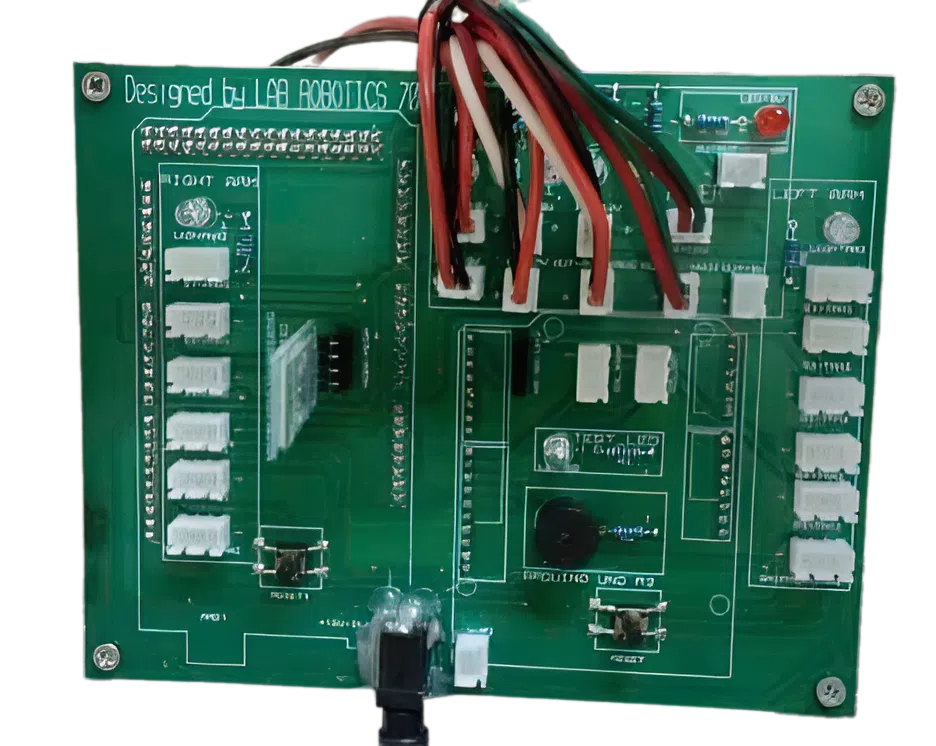}
\end{subfigure}
\caption{Simulation and real-world PCB implemented on the robot}
\label{fig:pcbonrobot}
\end{figure}

To enhance expressiveness and human-like communication, the robot’s head is equipped with 2 DOFs (yaw and pitch) for directional attention and expressiveness. A 7 in LCD screen mounted on the face region is used to display facial animations corresponding to the robot's emotional state. Additionally, a ZED 2 stereo camera is mounted above the screen to provide RGB and depth perception for object detection and gesture recognition. All the electronic components are coordinated through a custom-designed controller board, which serves as the central backbone of the hardware system. The PCB measuring $12 \times 12$ cm was developed to reduce wiring complexity, improve reliability, and provide a unified communication interface for all modules. A dual-microcontroller architecture is employed: an Arduino Mega 2560 is dedicated to controlling the 12 DOF of the dual arms, chosen for its large number of I/O pins, while an Arduino Uno manages the head module and sensors.  Both microcontrollers communicate via the I$^2$C protocol, ensuring synchronized signal exchange with minimal wiring complexity. The controller board also integrates a Bluetooth module, providing a wireless channel for lightweight teleoperation and simple experimental tasks, such as joint testing, trajectory playback, or app-based interaction. Fig.~\ref{fig:pcbonrobot} illustrates both the simulated PCB layout and the real-world PCB mounted on the robot.

The NVIDIA Jetson AGX Xavier serves as the main processing unit, running ROS for system communication and coordination. It also handles AI tasks, including computer vision, gesture classification, natural language processing, and trajectory planning for manipulation.
The Jetson is integrated with a custom controller board to form a scalable architecture: microcontrollers handle low-level motor control, while the GPU module manages high-level perception, planning, and decision-making. This layered design separates real-time control from AI computation, enabling both responsive performance and intelligent behavior in a compact, cost-effective humanoid platform.
\begin{figure}[!ht]
\centering
\includegraphics[width=0.5\textwidth]{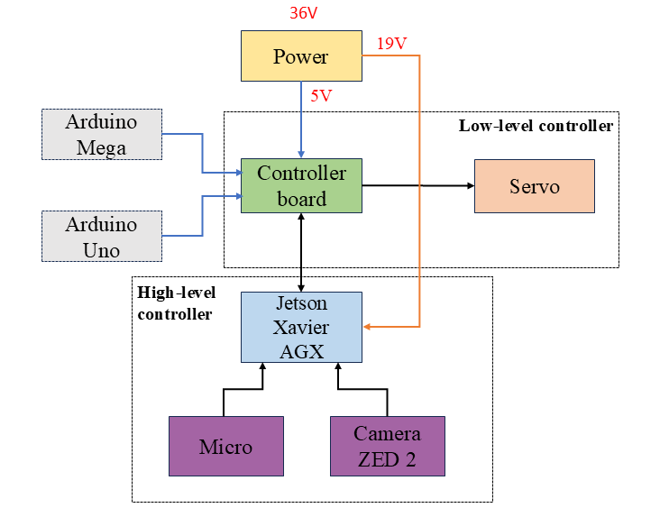}
\caption{The components in our robot}
\label{fig:components}
\end{figure}

The power and communication system of the humanoid robot is shown in Fig.~\ref{fig:components}. A main ${36}{V}$ supply provides power, which is stepped down to ${19}{V}$ for the Jetson processor and to ${5}{V}$ for the custom controller board modules. The ZED 2 stereo camera and microphone are connected directly to the Jetson. All servos run on the regulated ${5}{V}$ line, with their control signals managed through a dedicated controller board. This layered setup guarantees stable power distribution, organized wiring, and a clear separation between computing, sensing, and actuation. The robot's components and their key technical specifications are summarized in Table {\ref{tab:specific}}.
\begin{table}[!ht]
\centering
\caption{Technical specification of components}  
\label{tab:specific}
\begin{tabularx}{\linewidth}{  | >{\centering\arraybackslash}m{0.09\textwidth} 
  | >{\centering\arraybackslash}m{0.255\textwidth} 
  | >{\centering\arraybackslash}X|  } 
\toprule
\textbf{Device} & \textbf{Parameters} & \textbf{Application} \\ \hline

Battery & Input: 220V AC; Output: 36V 10Ah. & Power Supply \\ \hline

Jetson Xavier AGX & CPU: 8-core ARM v8.2; GPU: 512-core Volta Tensor Core; RAM: 32 GB LPDDR4x; Storage: 32 GB eMMC; Power: 10–30W. & High-level controller \\ \hline

Arduino UNO & MCU: ATmega328P; $14$ digital I/O; $6$ analog input; UART/I2C/SPI; $16$ MHz & Low-level controller \\ \hline

Arduino Mega 2560 & MCU: ATmega2560; $54$ digital I/O; $16$ analog input; $4$ UART; $16$ MHz & Low-level controller \\ \hline

FSR402 Force Sensor & Force range: $0.2$–$20$N; Resistance range: $\sim1.8$ k$\Omega$–10M$\Omega$; Diameter: $18$ mm &  Grasp feedback \\ \hline

ZED 2 Stereo Camera & Dual $4$ MP; FoV: $110^{\circ}$; Depth sensing: $0.2$-$20$ m; Integrated IMU + Barometer sensor; USB $3.0$ & Vision \\ \hline

Microphone & USB $3.5$ mm jack; Frequency response: $20$–$20$ kHz & Voice acquisition \\ \hline

Touch LCD & Resolution: $1024 \times 600$; Size: $7$ in; IPS panel; Capacitive touch; Interface: HDMI & Display interface \\ \hline

Servo & Detail in Table {\ref{tab:servo_specs}} & Joint actuator \\ 
\bottomrule
\end{tabularx}
\end{table}

\begin{figure*}[!ht]
\centering
\includegraphics[width=0.6\textwidth]{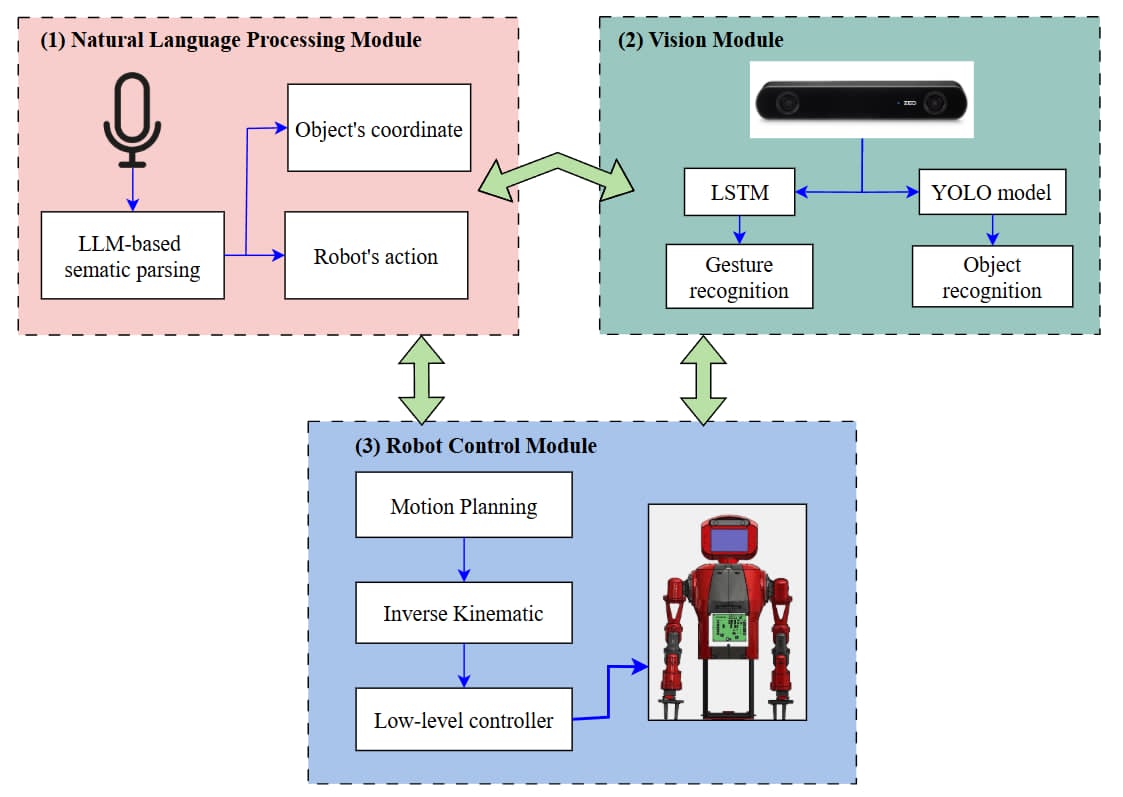}
\caption{Overview of the proposed software architecture}
\label{fig:overview}
\end{figure*}

\subsection{AI Integration and Software Architecture}

Fig.~\ref{fig:overview} illustrates an overview of our software system, which is organized into three modular AI modules that collectively enable the robot to perceive, understand, and respond to human interaction cues: (1) object detection, (2) gesture recognition, and (3) voice-command interpretation.

\subsubsection{Object recognition}

The object detection module used the ZED2 stereo camera to acquire RGB and depth data from the environment. A YOLO-based object detector~\cite{wang2024yolov9} is applied to the RGB stream to identify object classes and bounding box in pixel space. These pixel coordinates are then projected into 3D world coordinates using the depth map provided by the stereo camera. To ensure accurate localization, intrinsic calibration of the stereo camera is performed using a standard chessboard pattern to correct for lens distortion. The final $(X_r, Y_r, Z_r)$ coordinates are used to compute inverse kinematics and generate control commands for the robot arms, enabling precise object manipulation within the workspace.

\subsubsection{Gesture recognition}
The gesture recognition module facilitates non-verbal communication. It consists of three stages: human pose estimation, temporal feature extraction, and classification. MediaPipe Pose~\cite{googleaiedge_mediapipe} is employed for real-time 2D keypoint extraction, while a Long Short-Term Memory (LSTM)~\cite{hochreiter1997long} network is used to classify sequences into predefined gesture categories such as waving, handshaking, and love gestures. The model was trained on a custom dataset of gesture sequences under varied lighting and distance conditions. Once a gesture is detected, the robot responds with corresponding physical motion and facial expressions displayed on the LCD screen. This module enables the robot to engage in intuitive non-verbal communication with users, an essential capability for social HRI.

\subsubsection{Voice Command Understanding}

The voice interaction module allows the robot to receive and interpret spoken commands. Voice input is captured via a microphone and transcribed using the SpeechRecognition~\cite{uberi_speech_recognition}. The transcribed text is then passed to an LLM via a structured prompt for semantic parsing. The model outputs an action-object-location triplet in structured JSON format. For example, the voice command from the user “Take a cup and put it on the plate” is interpreted as:

\begin{Verbatim}[frame=single, numbers=left, numbersep=5pt, fontsize=\small]
{
    "action": "pick", "place"
    "objects": "cup,
    "destination": "plate"
}
\end{Verbatim}

If the objects and locations are detected by the vision module, their 3D coordinates are retrieved, and the instruction is translated into robot commands such as:
    \begin{Verbatim}[frame=single, numbers=left, numbersep=5pt, fontsize=\small]
     PICK(x1,y1,z1), MOVE(x2,y2,z2)
\end{Verbatim}

These are forwarded to the motor control unit to perform the intended task.

\section{Experimental Results} \label{sec:results}
To validate the effectiveness of the proposed humanoid robot system in real-world human–robot interaction (HRI) tasks, two experimental scenarios were implemented: (1) gesture-based interaction and (2) voice-command execution. All tests were conducted in an indoor laboratory environment under consistent lighting and minimal occlusion to ensure stable sensing performance. Videos demonstrating the results are available at: \url{https://youtu.be/NVSn821oaCY}

\subsection{Experimental setup}
\label{subsec : exset}
The robot prototype was placed on a fixed base in a standing posture. A ZED2 stereo camera mounted on the robot's head was used for visual perception, while a directional microphone was positioned near the user for voice input. Each experiment was repeated multiple times to evaluate response accuracy, latency, and robustness.
\begin{figure}[!ht]
\centering
    \begin{subfigure}[b]{0.38\textwidth}
    \includegraphics[width=\textwidth]{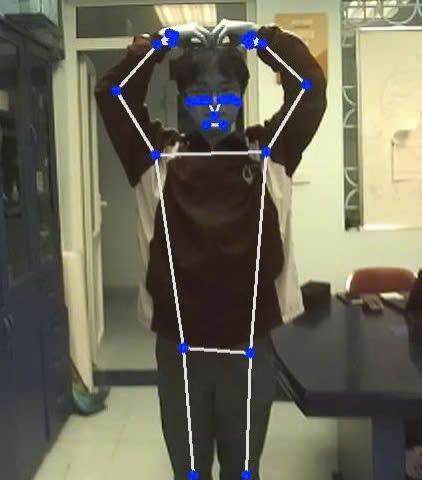}
    \caption{Love action recognition}
    \end{subfigure}
    \hfill
    \begin{subfigure}[b]{0.38\textwidth}
    \includegraphics[width=\textwidth]{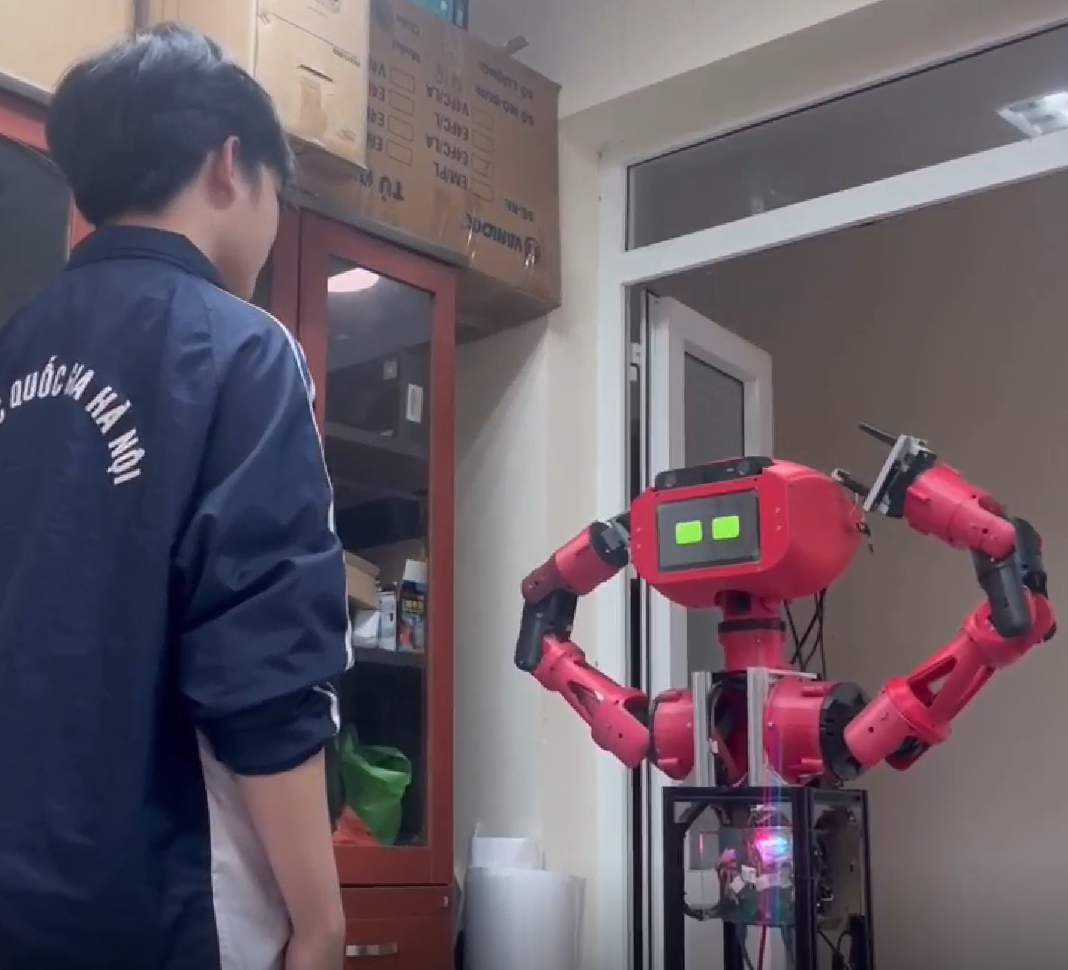}
    \caption{Robot action}
    \end{subfigure}
\caption{Illustration of human-robot interaction when recognizing love gesture action.}
\label{fig:actionrecog}
\end{figure}

\subsubsection{Scenario 1: Gesture-based Interaction}
The robot faced the user from a distance of approximately $1.5 m$. The gesture recognition module processed real-time skeletal data extracted by MediaPipe Pose and classified sequences using the trained LSTM model. Detected gestures, waving, loving, or handshaking, triggered predefined arm motions and corresponding emotional expressions displayed on the LCD screen.

\subsubsection{Scenario 2: Voice-command Execution and Object Detection}
In this scenario, the user issued verbal commands such as \textit{``Take the cup and put it on the plate''}. The voice was converted to text using speech recognition and parsed into executable robot instructions by an LLM. Simultaneously, the ZED2 stereo camera detected and localized the specified objects in 3D space using YOLO. The final robot command sequence was translated into pick-and-place operations using inverse kinematics.

\subsection{Results and Evaluation}

In this section, each AI module was evaluated independently and in combination to assess the robot's ability to respond accurately and naturally in HRI scenarios.

\subsubsection{Gesture Recognition Performance}
The LSTM-based gesture recognition module achieved a recognition accuracy of $96\%$ across three predefined gestures. The average response latency, measured from gesture completion to robot action initiation, was $1.2$ seconds. Example response, include waving back with a smiling face and audio playback, or performing a handshake when the corresponding gesture is detected. These behaviors demonstrate the robot’s ability to engage in non-verbal social cues, as illustrated in Fig.~\ref{fig:actionrecog}. 

\subsubsection{Voice Command and Object Detection Accuracy}
\begin{figure}[h!]
    \centering
    \begin{subfigure}[b]{0.24\textwidth}
        \includegraphics[width=\textwidth]{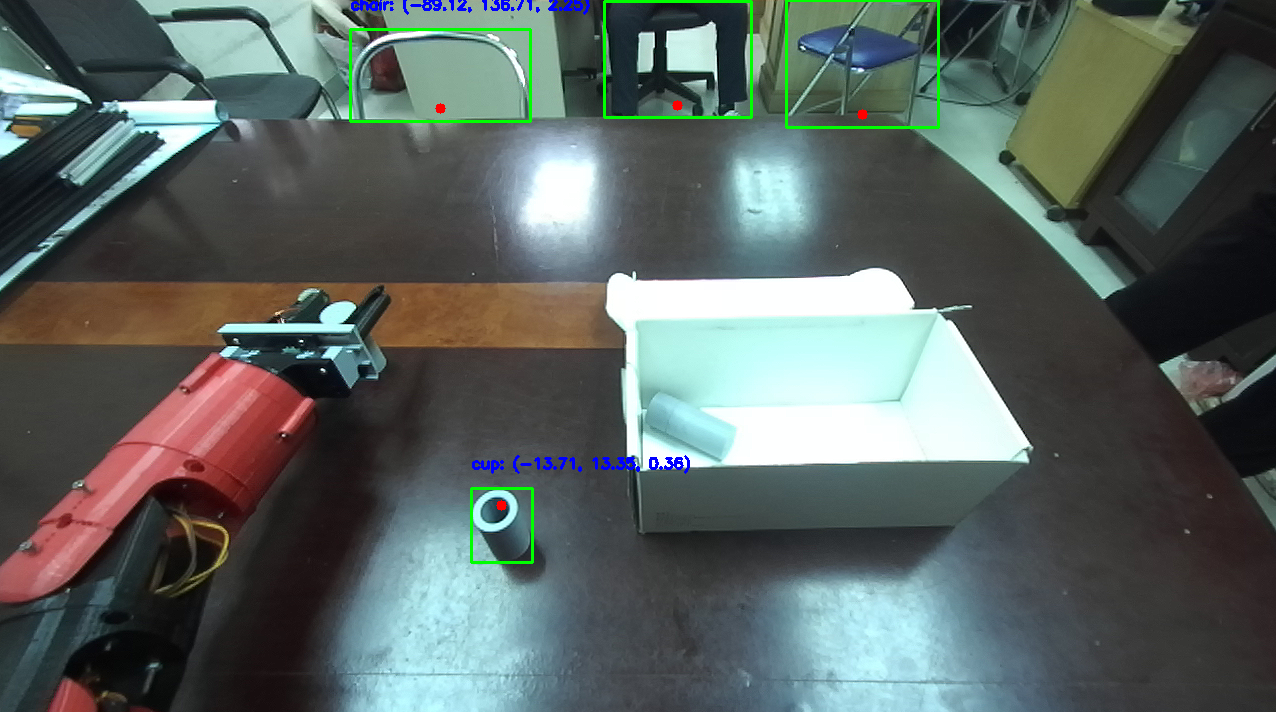}
        \caption{}
    \end{subfigure}
    \hfill
    \begin{subfigure}[b]{0.24\textwidth}
        \includegraphics[width=\textwidth]{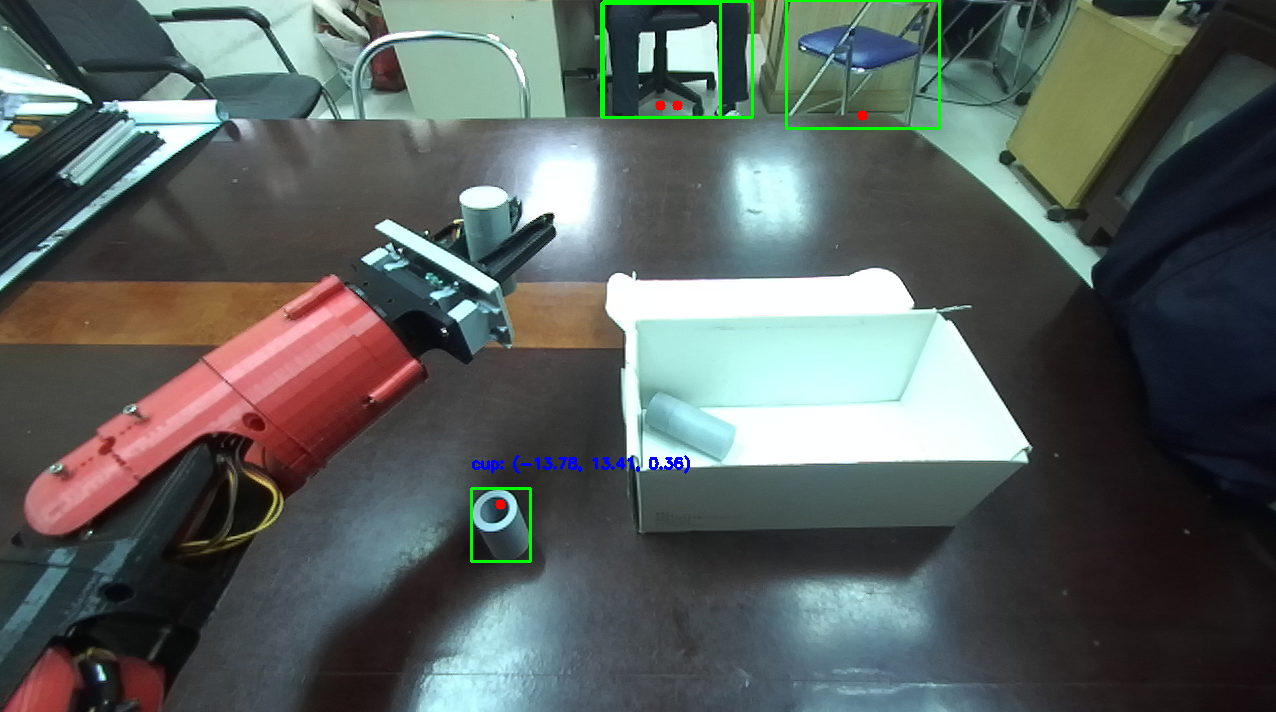}
        \caption{}
    \end{subfigure}

    \begin{subfigure}[b]{0.24\textwidth}
        \includegraphics[width=\textwidth]{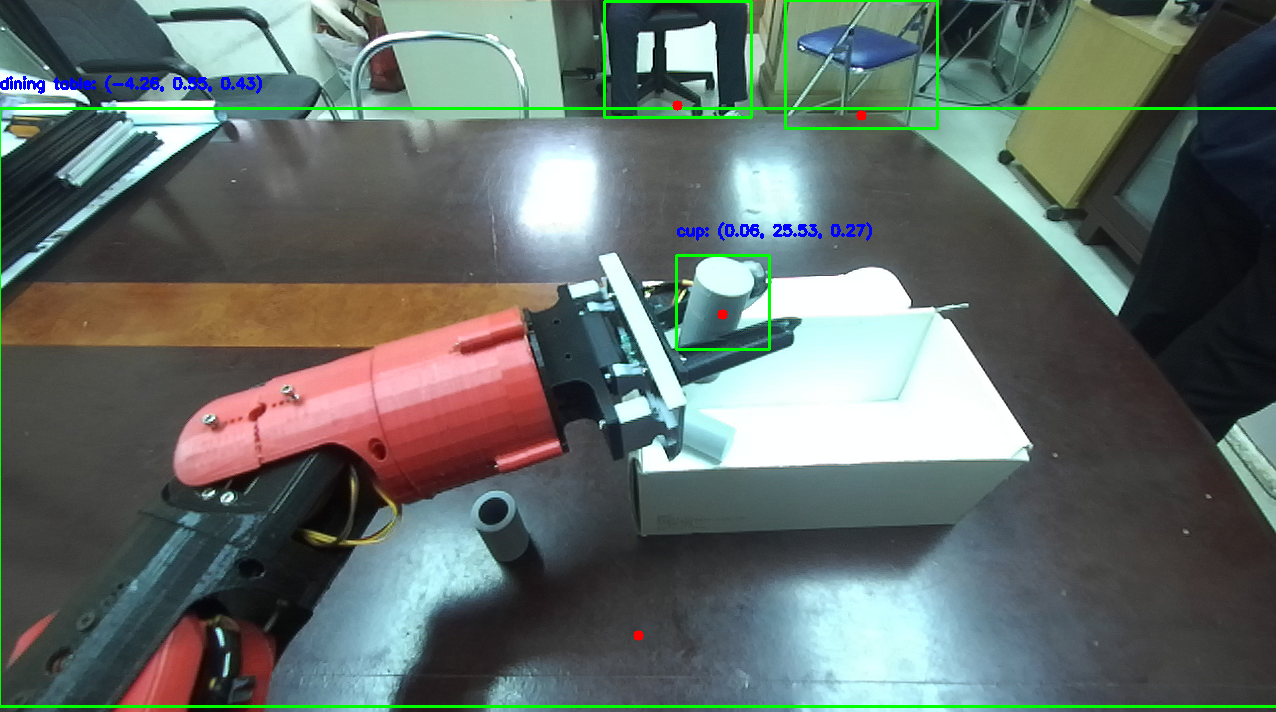}
        \caption{}
    \end{subfigure}
    \hfill
    \begin{subfigure}[b]{0.24\textwidth}
        \includegraphics[width=\textwidth]{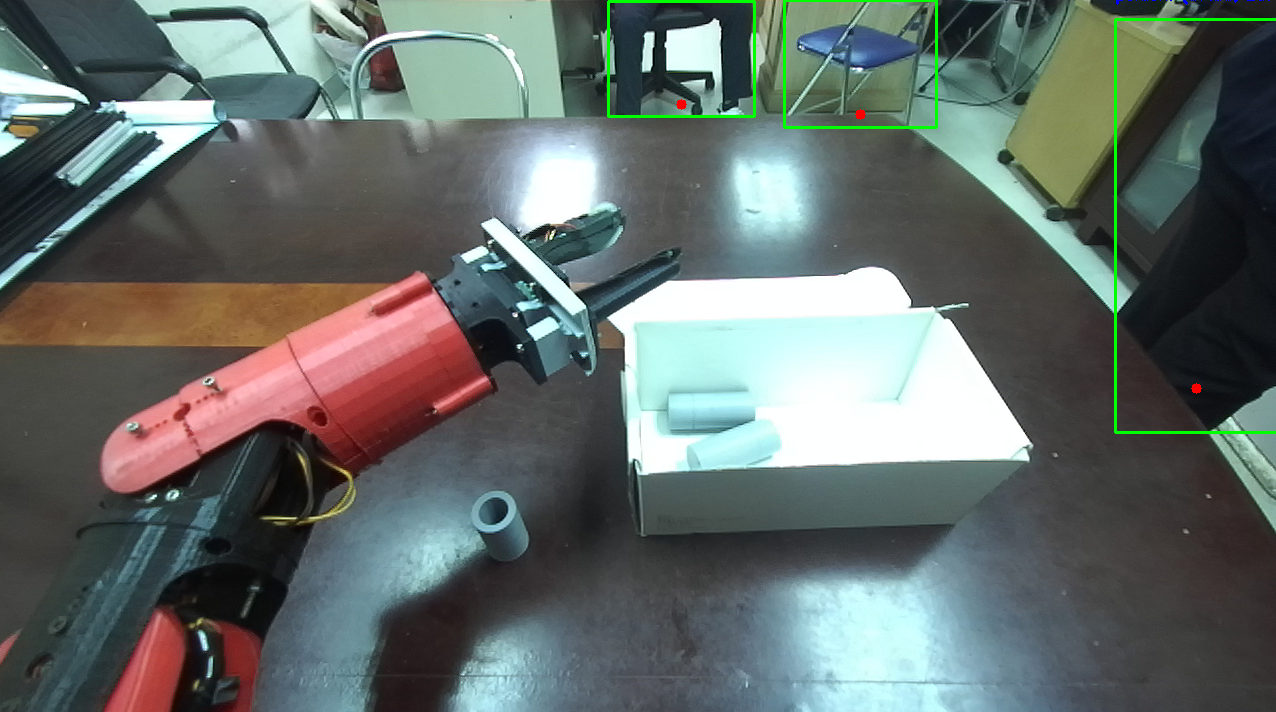}
        \caption{}
    \end{subfigure}

    \caption{Robot recognizing and interacting with objects.}
    \label{fig:robot-put-object}
\end{figure}

The voice-command pipeline was evaluated using $300$ randomly issued spoken instructions. The speech recognition system achieved a $92\%$ transcription accuracy, while the LLM-based intent parser achieved $96\%$ semantic parsing accuracy. YOLO-based object detection maintained over $90\%$ accuracy across varying lighting conditions and object types. Once the commands were parsed and the objects detected, the robot successfully executed the pick-and-place tasks. Despite relying on the Gemini API, the system maintained a short response time, generating actions within about 2 seconds on average. This efficiency was achieved because representing the output as structured JSON commands required fewer tokens to be generated, thereby reducing processing latency. As a result, the system satisfied the requirements for real-time scenarios. An illustration of the robot recognizing and interacting with objects is shown in Fig.~\ref{fig:robot-put-object}.

\subsubsection{Positioning and Manipulation Precision:}
The robot’s manipulation accuracy was evaluated by measuring the Euclidean error between target and actual positions across 10 trials. The average positional errors are summarized in Table~\ref{tab:error-results}.

\begin{table}[!ht]
\centering
\caption{Results of performance evaluation in position error of each arm (cm)}
\label{tab:error-results}
\begin{tabularx}{\linewidth}{  | >{\centering\arraybackslash}m{0.05\textwidth} 
  | >{\centering\arraybackslash}m{0.1\textwidth} 
  | >{\centering\arraybackslash}m{0.1\textwidth}  
  | >{\centering\arraybackslash}X|  }
\toprule
\textbf{Times} & \textbf{Left arm} & \textbf{Right arm} & \textbf{Integrated system} \\
\midrule
1  & 1.05 & 1.07 & 1.30 \\
2  & 1.28 & 0.88 & 1.24 \\
3  & 1.29 & 0.96 & 1.84 \\
4  & 1.55 & 1.36 & 3.48 \\
5  & 1.56 & 1.68 & 3.82 \\
6  & 1.25 & 1.24 & 1.51 \\
7  & 0.88 & 1.87 & 0.41 \\
8  & 1.86 & 0.88 & 0.97 \\
9  & 0.78 & 1.50 & 1.50 \\
10 & 1.48 & 1.40 & 2.26 \\
\midrule
\textbf{Average} & \textbf{1.30} & \textbf{1.28} & \textbf{1.83} \\ \bottomrule
\end{tabularx}
\end{table}

The system demonstrates stable performance in localization and manipulation tasks. The robot can reach and manipulate target objects within an average positioning error of approximately $1.3$ cm for both arms. Integration with YOLO introduces additional errors due to detection and transformation noise, with an average error of $1.83$ cm. Overall, the robot achieves a pick-and-place success rate of approximately $90\%$, confirming the reliability of the combined perception and control system in real-world HRI scenarios.

\section{Conclusion} \label{sec:conclusion}
This paper presented the design, development, and evaluation of a humanoid robot prototype for real-world human-robot interaction tasks. The system integrates a 12-DOF dual-arm mechanism, a 2-DOF expressive head with an LCD screen, and a custom-designed control board, forming a modular and scalable hardware platform. On the software side, the robot incorporates three core AI modules: gesture recognition using MediaPipe Pose and an LSTM network, object detection with YOLO and stereo-based 3D localization, and natural language understanding through speech recognition and an LLM-based semantic parsing. Two real-world interaction scenarios, gesture-based communication and voice command-driven object manipulation, were implemented and tested. Experimental results demonstrate performance across all modules, with gesture recognition achieving up to $96\%$ accuracy, voice command execution approximately $90\%$ successful, and pick-and-place operations maintaining an average positioning error of less than $1.9$ cm. Despite the system’s strong performance, a key limitation lies in its reliance on external LLM APIs, which introduce latency and dependency on internet connectivity. Future work will focus on integrating lightweight, on-device language models to improve real-time responsiveness.

\section*{Acknowledge} 
This work was supported by the JST SPRING, Japan Grant Number JPMJSP2102.

\nocite{*}
\bibliographystyle{IEEEtran}
\bibliography{references}

\end{document}